\documentclass[journal]{IEEEtran}

\usepackage{cite}
\usepackage[pdftex]{graphicx}
\graphicspath{{../img/}}
\usepackage[caption=false,font=footnotesize]{subfig}
\usepackage{siunitx}
\usepackage{amsmath}
\usepackage{amssymb}
\usepackage{multirow}
\usepackage{tabularx}
\usepackage[T1]{fontenc}
\usepackage[ruled,vlined]{algorithm2e}
\usepackage[export]{adjustbox}
\usepackage{enumitem}
\usepackage{subfig}
\usepackage{lipsum}
\usepackage[hyphens]{url}
\usepackage[scr]{rsfso}
\usepackage{tikz}
\usepackage{hyperref}

\newcommand\copyrighttext{\footnotesize \textcopyright~2022 IEEE. Personal use of this material is permitted. Permission from IEEE must be obtained for all other uses, in any current or future media, including reprinting/republishing this material for advertising or promotional purposes, creating new collective works, for resale or redistribution to servers or lists, or reuse of any copyrighted component of this work in other works.
DOI: \href{https://doi.org/10.1109/TIV.2022.3164236}{10.1109/TIV.2022.3164236}%
}

\newcommand\copyrightnotice{%
    \begin{tikzpicture}[remember picture,overlay]%
 	\node[anchor=south, xshift=0pt, yshift=4pt] at (current page.south)%
 	{\fbox{\parbox{\dimexpr\textwidth-\fboxsep-\fboxrule\relax}{\copyrighttext}}};%
 	\end{tikzpicture}%
}
\begin{document}

\title{Situation-Aware Environment Perception Using a Multi-Layer Attention Map}

\author{Matti~Henning,
        Johannes~Müller,
        Fabian~Gies,
        Michael~Buchholz
        and~Klaus~Dietmayer
\thanks{M. Henning, J. Müller, M. Buchholz and K. Dietmayer are with the Institute of Measurement, Control and Microtechnology of the University of Ulm, 89081, Ulm, Germany. E-Mail: <firstname>.<lastname>@uni-ulm.de}
\thanks{This research is accomplished within the UNICAR\emph{agil} project~\cite{Woopen2020} (FKZ
16EMO0290). We acknowledge the financial support for the project by the
German Federal Ministry of Education and Research (BMBF).}}


\maketitle

\copyrightnotice%
\begin{abstract}
Within the field of automated driving, a clear trend in environment perception tends towards more sensors, higher redundancy, and overall increase in computational power. 
This is mainly driven by the paradigm to perceive the entire environment as best as possible at all times. 
However, due to the ongoing rise in functional complexity, compromises have to be considered to ensure real-time capabilities of the perception system. 

In this work, we introduce a concept for situation-aware environment perception to control the resource allocation towards processing relevant areas within the data as well as towards employing only a subset of functional modules for environment perception, if sufficient for the current driving task. 
Specifically, we propose to evaluate the context of an automated vehicle to derive a multi-layer attention map (MLAM) that defines relevant areas. 
Using this MLAM, the optimum of active functional modules is dynamically configured and intra-module processing of only relevant data is enforced.

We outline the feasibility of application of our concept using real-world data in a straight-forward implementation for our system at hand.
While retaining overall functionality, we achieve a reduction of accumulated processing time of 59\,\%. 
\end{abstract}%

\IEEEpeerreviewmaketitle

\section{Introduction}
\label{sec:intro}
\IEEEPARstart{S}{ince} the start of the century, the field of Advanced Driving Assistance Systems (ADAS) in the automotive area has grown both in popularity as well as in availability on the consumer market. 
Vehicles equipped with these systems rely on cost-efficient automotive sensors and provide functionality with respect to SAE level 1 or 2~\cite{SAE_J3016_202104}, while SAE level 3 functionality has been announced.

In parallel, the research towards SAE level 4 and 5 has received significant attention. 
For these automated driving features, the trend with respect to environment perception generally tends towards the usage of highly sophisticated, multi-modal, and multi-redundant sensor systems~\cite{Liu2021}. 
This is driven by the paradigm to achieve the best possible environment perception at all times and for \SI[mode=text]{360}{\degree} around the vehicle.

\begin{figure}[t!]
    \centering
    \subfloat[Concept example of an MLAM constructed by the aggregation of three layers \textbf{L}]{
        \includegraphics[width=0.48\linewidth]{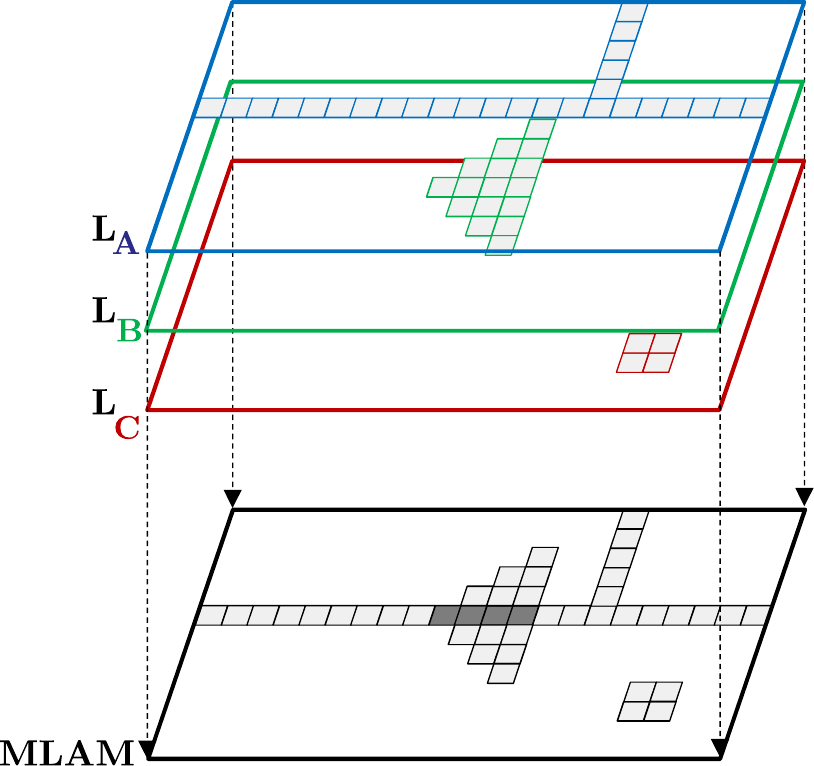}
        \label{fig:intro:layerExample:concept}}
    \hfil
    \subfloat[Real-world example of an MLAM with ego vehicle in blue and oncoming object in red.]{
        \includegraphics[width=0.4\linewidth]{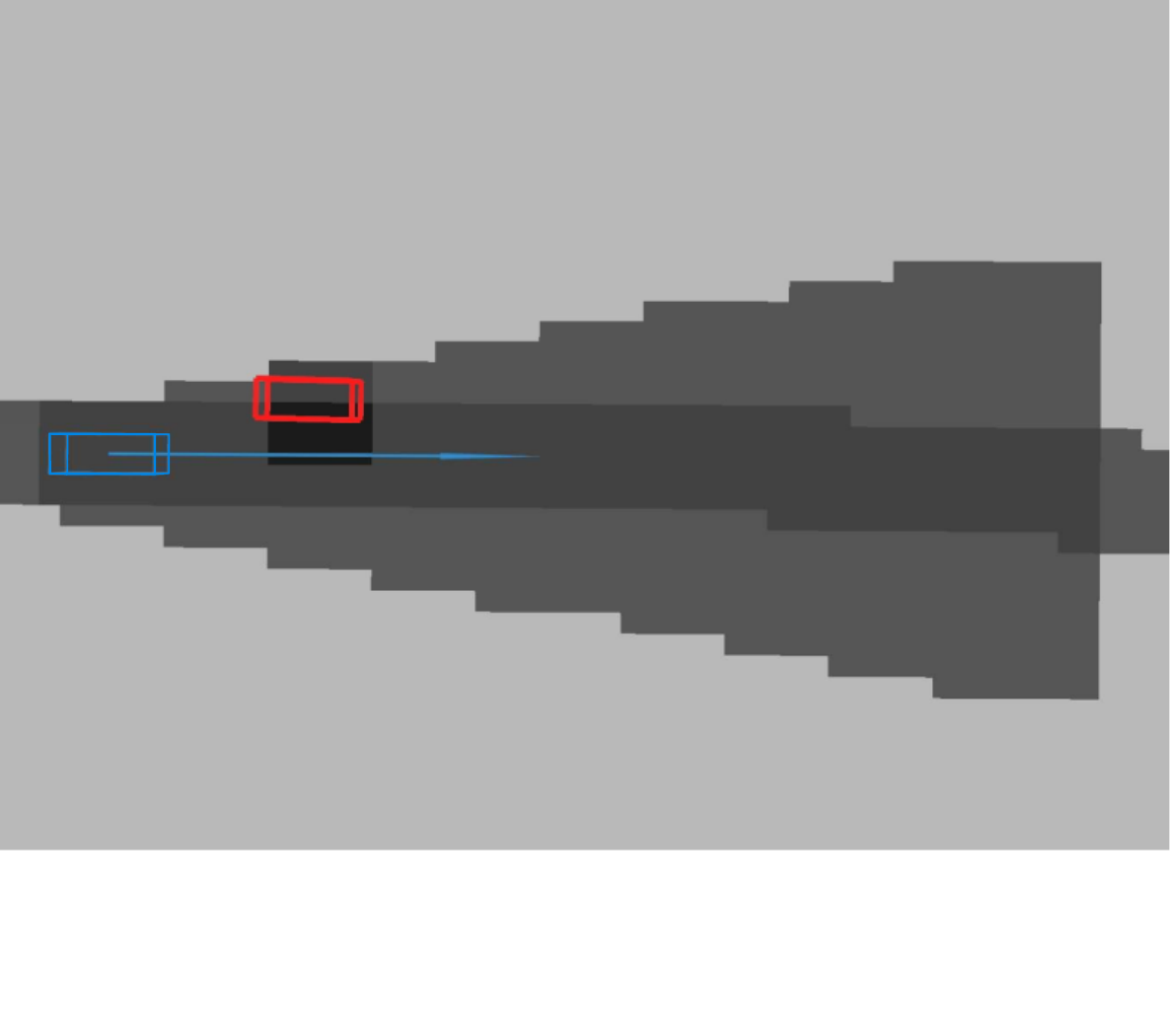}
        \label{fig:intro:layerExample:realworld}}
    \caption{Example for a Cartesian Multi-Layer Attention Map (MLAM) representation. Darker cells correspond to higher required attention.}
    \label{fig:intro:layerExample}
\end{figure}

Accordingly, cost-intensive state of the art processing systems are employed, that require a considerable amount of energy to operate at full capacity~\cite{Liu2021}, which contradicts the general goal of energy conservation and efficient vehicles.
Still, the available computational resources are often insufficient~\cite{Liu2021} and compromises have to be made to ensure real-time capabilities.

One straight-forward possibility is to reduce the computational requirements by focusing only on relevant data, i.e. reducing the overall amount of processed data. This is often referred to as applying \textit{attention} to relevant data.  
Within the last decade, several variants of this approach have been published.
Approaches using saliency, e.g. ~\cite{Bajcsy2018, Walsman2019, Rasouli2020, Pal2020}, are based on the assumption that objects that are closer to the perceiving vehicle are more relevant than objects that are further away.
Other approaches, like~\cite{Xu2015, Homeier2011, Kootbally2008, Nager2021}, relate relevance of data to a-priori knowledge of the context, e.g., derived from the \textit{situation}. 
They employ simple masks to filter out non-relevant data in an heuristic manner.
Figure~\ref{fig:intro:layerExample} provides an example for such masks, if interpret binary between light and dark cells. 

All these examples share the key idea to first perceive the entire environment on sensor level and then apply some sort of filtering before further processing. 
This approach emerged recently due to the increasing availability of cost-efficient sensor and processing hardware.
Before this increase, research in the field of \textit{active perception} employed actuated sensors or sensor platforms to actively control gazing direction, e.g. \cite{Bajcsy2018, Pellkofer2000, Seo2008, Unterholzner2012}.
Defining the gazing direction over time is commonly represented by an optimization function that needs to be maximized. 
Often, information maximization, entropy, or similar notations are used. 
While actuated sensor platforms have been replaced in current automated vehicles, the concepts and solutions from \textit{active perception} become more relevant for state-of-the-art systems. Bajcsy et al.~\cite{Bajcsy2018}, who have reviewed the progress of the past decades in the field of active perception a few years ago, draw this conclusion as well. 

The term \textit{active perception} is closely related to the term \textit{situation-awareness}.
Occasionally they are used interchangeably. 
For the purpose of our work, we use situation-awareness as a broader term, that includes active perception mechanisms as well as attention mechanisms. 
Our definition of the \textit{situation} is based on Ulbrich et al. \cite{Ulbrich2015}.

\subsection*{Contribution}
In this work, we present a scalable and modular concept for situation-aware environment perception, that, to the best of our knowledge, is unprecedented in the current research.

In Section~\ref{sec:concept}, we first introduce a general formulation to identify relevant data and how to focus the perception processing chain accordingly. 
Then, we describe our concept in three steps:
\begin{enumerate}
    \item High-level situation analysis is linked to a set of relevance defining layers. 
    The layers are combined into a multi-layered attention map (MLAM, cf. \figurename~\ref{fig:intro:layerExample}), that defines requirements for the perception processing chain. 
    \item The MLAM is cross-checked against a module configuration tree to find the best configuration. The processing chain is reconfigured during runtime. 
    \item The MLAM is applied within the active modules. 
\end{enumerate}
Compared to the the all-time \SI[mode=text]{360}{\degree} perception, the key benefit of our concept is the reduction of resource consumption. This is achieved by reducing the average number of computations to be executed as well as by efficient allocation of available resources instead of uniform resource distribution.

We show the viability of our concept by applying it to our automated driving system at hand in a straight-forward manner in Section~\ref{sec:proof}.
Providing a detailed concept, we aspire to start an engaging discussion about implementation variants and future possibilities for situation-aware environment perception.

\section{Related Work}
\label{sec:rw}
For situation-awareness, individual solutions for specific setups and/or problems exist, where resources are allocated dynamically towards relevant data (cf. examples from Section~\ref{sec:intro}). High-level concepts for architectures that enable the application of these solutions in a more generic way exist as well, although they do not provide details on their implementation.  
Albus~\cite{Albus1999, Albus2002} introduced a layered architecture for task planning of autonomous agents in 1999, including modules for the inclusion of a-priori knowledge. 
Gehrke~\cite{Gehrke2008} introduced requirements a situation-aware system must meet that are derived from the levels of situation-awareness from Endsley~\cite{Endsley1995}.
To the best of our knowledge, only one approach provides actual implementation details and performance results: 
Matzka et al. \cite{Matzka2012}, relating to Matzkas thesis from 2009~\cite{Matzka2009}, introduced a situation-aware system for a forward-looking sensor setup that focuses on highway scenarios. 
Regions of interest, corresponding to patches of an image, are defined using a-priori knowledge and online information. 
These regions are further investigated based on their \textit{utility}, and a subset of available modules for traffic participant detection and classification is deployed for the most useful regions. 
However, the interpretation of relevant data as a set of image patches restricts the applicability of their system for a generic processing chain for automated driving as per Ulbrich et al.~\cite{Ulbrich2017}. 
Similarly, their system does not generalize as a concept for any type of automated vehicle, but is instead tailored to the vehicle they have used. 
These design decisions were reasonable at that time, but do no longer apply to the state of the art in automated driving. 
Employed sensors, sensor coverage of the environment, and the required flexibility in driving tasks to be executed require a flexible and scalable concept that ideally applies to all automated vehicles. 

To implement such a flexible concept, the software architecture shall support dynamic connections between modules and/or dynamically changing interfaces. 
The same holds for the other way around, The same holds for the other way around, as architectures that support dynamic reconfiguring need to decide what to configure.
Available architectures are, among others, SAFER developed by Kim et al.\cite{Kim2012} using standby modules to reconfigure in case of a module failure or ASOA, introduced by Kampmann et al.~\cite{Kampmann2019} using an \textit{orchestrator} to reconfigure the active software modules in real time during operation. 
Schlatow et al.~\cite{Schlatow2015} formulate a more general, constraint based approach. 
Also, ROS2~\cite{Thomas2014, ros22021} supports reconfiguring using managed life-cycle nodes.
With our work on a scalable and modular frame for situation-aware environment perception, we aim to support configuration decisions. 
However, details on software architecture are out of the scope of this work.

\section{Situation-aware Environment Perception}
\label{sec:concept}
In this section, we first introduce a general formulation of the problem to be solved and the corresponding assumptions to solve it. Then, our solution is introduced in three steps:
\begin{enumerate}
    \item situation detection and attention map generation,
    \item processing chain configuration, and 
    \item attention map application.
\end{enumerate}
We name the application of these steps \textit{awareness processing}.

\subsection{Problem Formulation}
\label{sec:concept:problem}
Our formulation of a systematic benefit of awareness processing is based on the assumption that, for any situation $\mathbf{s} \in \mathbb{S}$, the surrounding environment $\mathbb{E}$ of an automated vehicle is separable into \textit{relevant} and \textit{non-relevant} regions. 
A region corresponds to an area around the vehicle and can be represented by any geometric description, e.g., a cell in a Cartesian grid or closed polygons.
Additionally, for every region, a quantifiable value of relevance in comparison to other regions can be set.

To derive a separation of the environment $\mathbb{E} = \{r_i \}_{i=0 ... \mathbf{N}}$, which contains the regions $r_i$, into two subsets of \textit{relevant environment} $\mathbb{E}_r$ and \textit{non-relevant environment} $\mathbb{E}_n$, we introduce the \textit{relevance operator}
\begin{equation}
    \text{rel} \left\{ \cdot \right\} : \mathbb{E} \times \mathbb{S} \mapsto \mathbb{R}^{+}_0.
\end{equation}
We can assign appropriate relevance to regions of the environment depending on the current situation $\mathbf{s}$:
\begin{subequations}
    \label{eq:relEnv}
    \begin{align}
            \mathbb{E} &= \mathbb{E}_r \cup \mathbb{E}_n \, , \\
            \mathbb{E}_{r}\left(\mathbf{s}\right) &= \left\{ r_i \in \mathbb{E} : \text{rel}\left\{r_i, \mathbf{s} \right\} > \theta_{\text{rel}} \right\}.
    \end{align}
\end{subequations}
A region in the environment becomes relevant above a threshold $\theta_\text{rel}$, which might be $\theta_\text{rel} = 0$ for most applications.

With relevant regions identified, we interpret the perception processing chain as a set of functional modules $\mathbb{M}$. 
The optimal subset of active functional modules $\mathbb{M}^\ast$, selected from the available functional modules $\mathbb{M}$, is configured based on the relevant regions. 
To obtain $\mathbb{M}^\ast$, we introduce the configuration operator
\begin{subequations}
    \begin{align}
            \text{conf}\{ \cdot \}&: \mathbb{S} \times \mathscr{P}(\mathbb{E}) \mapsto \mathscr{P}(\mathbb{M}) \, , \\
            \mathbb{M}^\ast &= \text{conf}\left\{ \mathbf{s}, \mathbb{E}_r \right\}\text{,} \hspace{.25cm} \mathbb{M}^\ast \subseteq \mathbb{M}.
    \end{align}
\end{subequations}
The powerset operator $\mathscr{P}\{\cdot\}$ is used to generate the set of subsets of $\mathbb{E}$ and of $\mathbb{M}$, including the set itself as well as the empty set $\emptyset$.
The $\text{conf}\{ \cdot \}$ operator hence maps the current situation $\mathbf{s}$ and any combination of relevant regions $\mathbb{E}_r$ to the optimal combination of functional modules $\mathbb{M}^\ast$ out of all combination candidates $\mathbb{M}' \in \mathscr{P}\{ \mathbb{M} \}$.

\subsection{Assumptions}
\label{sec:problem:assumption}
The problem formulation and our following proposed solution is based on the following assumptions:
\begin{enumerate}
    \item A description for the environment $\mathbb{E} = \left\{ r_i \right\}_{i=0...N}$ exists, which enables for every region $r_i$ to set a quantifiable performance requirement $p_i^{\text{req}}$ using $\text{rel}\{\cdot\}$.
    \label{sec:concept:lemma:regions}
    \item The performance and cost of the modules $m \in \mathbb{M}$ can be quantified and their metrics are additive. 
    \label{sec:concept:lemma:pc}
    \item Any module $m$ relates to a subset of regions $\mathbb{E}_m \subseteq \mathbb{E}$, representing their coverage of the environment. 
    Furthermore, it is known which modules are data \textit{sources} and which modules are \textit{non-sources}.
    \label{sec:concept:lemma:dataSources}
    \item The linkability between modules is known. Thus, the processing chain candidates $\mathbb{M}'\subseteq \mathbb{M}$ can be constructed and evaluated for performance.
    \label{sec:concept:lemma:relations}
\end{enumerate}
Our approach is based on the hypothesis that the resources saved by awareness processing are, on average, larger than its computational overhead.   
We provide a verification of this hypothesis for our application in Section~\ref{sec:proof}.

\subsection{Situation Detection and Attention Map Generation}
\label{sec:concept:MLAM}
As a first step to find the optimal configuration for a perception system of an automated vehicle based on its situation, the influences of the situation that are relevant for the perception need to be described. 
The key idea of our proposed approach is that any situation $\mathbf{s} \in \mathbb{S}$, with $\mathbb{S}$ denoting all possible situations, is linked to a set of attention layers $\mathbb{L}$, as visualized in Table~\ref{tab:concept:layerLink}.
The layers define the relevance of regions using the relevance operator $\text{rel}\{\cdot\}$.
\begin{table}
\centering
\caption{Example for the activation of attention layers.}
\label{tab:concept:layerLink}
\begin{tabular}{lc|ccccc}
                        &        & Layer A                    & Layer B                    & Layer C                    & Layer D                    & ...\\
                        \hline
\multicolumn{1}{r}{$\mathbf{s}_i$} & \multirow{4}{*}{\rotatebox[origin=c]{90}{situation}} & x                    & x                    &                      &       &               \\
\multicolumn{1}{r}{$\mathbf{s}_j$} &                            &                      & x                    &                      & x                 &   \\
\multicolumn{1}{r}{$\mathbf{s}_k$} &                            &                      & x                    & x                    & x                 &   \\
...                     &                            & \multicolumn{1}{l}{} & \multicolumn{1}{l}{} & \multicolumn{1}{l}{} & \multicolumn{1}{l}{} &
\end{tabular}
\end{table}
Any situation $\mathbf{s}$ acts as activation function, so that a subset $\mathbb{L}_a \subseteq \mathbb{L}$ represents the active attention layers for the respective situation. This subset equals the empty set ($\mathbb{L}_a = \emptyset$), if no attention is required, e.g., in a parking or standby scenario.
Therefore, the set of possible situations $\mathbb{S}$ corresponds to a combination of high-level situational influences, e.g., the driving task or maneuver, the geographic location, or the weather conditions. 

The relevance of a region $r_i$ represents its performance requirement $p_i^\text{req}$ for the perception. Every attention layer represents an independent measure for relevance for the driving task. 
As a result, the indicated relevance values of multiple layers for the same region aggregate to the overall required performance 
\begin{equation}
    p_i^\text{req} \geq \max_{k=0...L}\left(p_{k,i}^{\text{req}}\right)\,,
\end{equation}
with $L$ representing the number of elements in $\mathbb{L}_a$ and $p_{k,i}^{\text{req}}$ being the assigned requirement for region $r_i$ by layer $k$.
Representing each layer as a function $\mathbf{l}(r_i)$ that assigns relevance to a region $r_i$, the aggregated performance requirement can be implemented as the sum of assigned relevance values from all active layers. 
The relevance operator $\text{rel}\left\{\cdot\right\}$ is then defined as 
\begin{equation}
\label{eq:concept:comb}
    \text{rel}\{r_i\} : p_{i}^{\text{req}} = \sum_{l_k \in \mathbb{L}_a} \mathbf{l}_k(r_i).
\end{equation} 

An example for our argumentation is to define two layers $x, y$ so that the corresponding function $\mathbf{l}_x$ assigns relevance to any region of driveable area for expected traffic participants and $\mathbf{l}_y$ assigns relevance to any region that intersects with the planned trajectory of the vehicle. 
Any traffic participant that is on a known expected area or on the planned trajectory should be perceived, so that the automation might react on its behavior. 
In the case that a traffic participant is both in a region where it is expected \textit{and} the region intersects with the planned trajectory, the quality of perception for this traffic participant should be higher. Consequently, the risk of a collision due to an incorrect perception should be lower.

Centering our description for relevant regions around the combination of layers, we label it \textit{multi-layer attention map} (MLAM). 
Figure~\ref{fig:intro:layerExample} provides a visual example describing the regions $r_i$ as cells in a Cartesian grid. 
In \figurename~\ref{fig:intro:layerExample:concept}, three active layers are combined into the MLAM.
Overlapping relevant regions result in a higher performance requirement that is visualized by the increasing darkness of the cell color. \figurename~\ref{fig:intro:layerExample:realworld} shows a similar example derived from real-world data. 
The oncoming object marked as red rectangle results in the region with highest performance requirement, again corresponding to the darkest cell color.

\subsection{Processing Chain Configuration}
\label{sec:concept:config}
Using the MLAM, the corresponding optimal subset of active modules $\mathbb{M}^\ast$ needs to be identified and configured. 
In this subsection, we introduce a module configuration tree and the optimization criterion to search for $\mathbb{M}^\ast$ within the tree.  

\subsubsection{Optimization Criterion}
Based on the outlined benefits of situation-awareness, we understand optimality as using the minimum of resources of the system $\mathcal{R}^\ast$  to achieve the required performance $\mathcal{P}^{\text{req}}=\left\{ p_i^{\text{req}}\right\}~\forall~r_i\in \mathbb{E}_r$ for relevant regions.
We therefore require that every module $m \in \mathbb{M}$ specifies its quantifiable \textit{performance} $p_m$ w.r.t. the task of automated driving and its quantifiable \textit{cost} $c_m$ in terms of resource consumption (see assumption \ref{sec:concept:lemma:pc}).

Consequently, the optimization criterion for awareness processing is given by:
\begin{align}
\mathbb{M}^\ast &=  \underset {\mathbb{M}' \subseteq \mathbb{M}} {\text{arg
min}} \left( 
\sum_{m \in \mathbb{M}'} c_m  \right)\label{eq:intro:RC_min}
~\text{s.t.}~\sum_{m \in \mathbb{M}'} p_m \geq \mathcal{P}^{\text{req}}.
\end{align}
Optimality might also be understood as using all available resources of the system $\mathcal{R}^{max}$ to achieve the highest performance for relevant regions $\mathcal{P}^{\ast}$, which is outside of the scope of this work.

As an example for the \textit{cost} of a module, in this work, we use the average module processing time, as this is an often encountered bottleneck. 
However, using further approaches to estimate the resource consumption, e.g., based on mathematical complexity, or using software profiling measures, e.g., tail latency, is also possible. 
Lin et al.~\cite{Lin2018} provide an example using tail latency for the evaluation of hardware acceleration benefits for automotive applications. 

To quantify the \textit{performance} of a module, metrics for several aspects of the environment perception in automated driving applications exist, e.g. OSPA~\cite{OSPA2008} or (generalized) IoU~\cite{giou2019} and $F_1$\textit{-Score} for object detection. 
As another example, Por\k{e}bski and Kogut~\cite{Porebski2021} introduced a metric for occupancy grid evaluation.
These metrics compare the performance of a module under test with the ground truth. 
As awareness processing alters where objects might be perceived, the identified relevant regions need to be considered in a similar manner for the ground truth for these metrics to be viable. 

Beside above mentioned metrics, a relative characterization of the modules performance can be derived from expert knowledge.
As this work focuses on the concept and its applicability, we choose this approach in Section~\ref{sec:proof}.

\subsubsection{Graceful Degradation and Self-Awareness}
The concept of awareness processing relates to the perception block of the functional system architecture in~\cite{Ulbrich2017}.
Planning and control modules are considered to be black-boxes that pose requirements for the perception block. 
These requirements are incorporated into the relevance operator $\text{rel} \left\{ \cdot \right\}$, e.g., by defining corresponding attention layers, and consequently influence the required performance $\mathcal{P}^{\text{req}}$.
If the required performance can be met by $\mathbb{M}^\ast$ derived from \eqref{eq:intro:RC_min}, the performance of the latter modules are assumed to be sufficient to solve the current automation task, while resource consumption is minimized. 
However, if the performance requirement from \eqref{eq:intro:RC_min} can not be met, the question arises if the continued execution of the current automation task is safe.
Strongly connected to this question is the research field of self-awareness, e.g., as described by Schlatow et al.~\cite{Schlatow2017}.
The idea of self-awareness is to evaluate both module and system level capabilities during runtime and derive decisions on how to act based on the current capabilities.

In the context of awareness processing, self-awareness can be used on system level to gracefully limit the functional range of the planning modules, e.g., by reducing the speed of the vehicle. 
Consequently, the corresponding performance requirement to the perception modules declines. 
If the performance requirement in \eqref{eq:intro:RC_min} can then be met, the automation, even with the limited functional range, is safe. 
Otherwise, emergency maneuvers or take-over functionalities might be initialized via self-awareness. 
Tas et al.~\cite{Tas2018} employ a system-wide performance monitoring that is centered around self-awareness. 
In the case of perception degradation, a performance assessment fusion module degrades the capabilities of the planning.
Another example is realized by the UNICAR\emph{agil} project~\cite{Woopen2020}, where self-awareness is used to enhance functional and operational safety of automated driving by monitoring provided and required capabilities~\cite{Stolte2020}.

Besides the degradation of system capabilities, from a module perspective, self-awareness provides means to measure the performance of and relations between modules during runtime. Our concept of situation-awareness for the perception system is designed to incorporate these dynamic changes in module performance using the module configuration tree, as described in the following. 
While we focus on situation-awareness in this work and do not address self-awareness in more detail, there is a strong potential for synergy between the two.

\subsubsection{Module Configuration Tree} 
Based on Ulbrich et al.~\cite{Ulbrich2017}, we classify the modules $m \in \mathbb{M}$ w.r.t. their algorithm class, e.g., object detection, semantic segmentation, or free space detection.  
A perception system for which awareness processing shall be applied might provide a handful of variants for these classes, so that the search space for the optimal configuration $\mathbb{M}^\ast$ remains manageable.
Considering a camera object detector as an example, numerous implementations exist~\cite{kitti3DList}, where variants might correspond to high-performing and high-cost, cost-effective or minimal-cost implementations.
In addition to class and variant, every module $m$ consists of the following properties:\\[4pt]
\begin{tabular}{rl}
    {type}         \hspace{.5cm} & $ \displaystyle t \in \left\{ \text{\textit{source}}, {\textit{non-source}} \right\} $ \\[2pt]
    {coverage}     \hspace{.5cm} & $ \displaystyle  r_j \in \mathbb{E}_m \subseteq \mathbb{E} $ \\[2pt]
    {performance}  \hspace{.5cm} & $ \displaystyle  p_j^{\text{prov}} ~\forall~ r_j \in  \mathbb{E}_{m} $\\[2pt]
    {cost}         \hspace{.5cm} & $ \displaystyle  c = f(\mathbb{E}_r) $\\[2pt]
    {relations}    \hspace{.5cm} & $ \displaystyle  R $
\end{tabular}\\[4pt]
The provided performance of a module $p_j^{\text{prov}}$ applies to every region that it covers (see assumption \ref{sec:concept:lemma:dataSources}). 
The cost of a module $c$ is represented as a function $f(\mathbb{E}_r)$ depending on the relevant regions. 
The cost function might be constant, e.g., for a neural network based algorithm processing constant amount of data per cycle, or might depend on the relevant area, e.g., for a tracking that grows in complexity with the number of object detections.

Linkabilities, dependencies on, or interactions between modules w.r.t. cost or performance are modeled by the relations $R$ (see assumption  \ref{sec:concept:lemma:relations}).
Here, strengths and/or weaknesses of specific module combinations can be considered. 
Besides, situation dependent relations can be modeled that dynamically change on a high level, e.g., a lane detection module must be included if the map is identified to be faulty. 
Another example is an object detection module that has only low performance under heavy weather conditions.
For the implementation of the relations, various possibilities exist.
In Section~\ref{sec:proof}, we present one option that uses white- and blacklists in the initialization phase as well as in the search for the optimal configuration during runtime.

Using the pre-defined module linkabilities as static constrains, a set of configuration trees can be constructed. 
The configuration trees only contain valid combination candidates $\mathbb{M}'$ and, thus, reduce the search space for the optimal configuration. Their nodes consists of:\\[4pt]
\begin{tabular}{rl}
    {configured modules}    & $ \displaystyle m \in \mathbb{M}' \subseteq \mathbb{M} $\\[3pt]
    {aggregated coverage}   & $ \displaystyle r_j \in \mathbb{E}_{\mathbb{M}'} = \mathop{\cup}_{m \in \mathbb{M}'} \mathbb{E}_m $\\[3pt]
    {aggregated performance} & $ \displaystyle \boldsymbol{p}_j^{\text{prov}} = \sum_{m \in \mathbb{M}'} p_{j,m}^{\text{prov}}   ~\forall~ r_j \in \mathbb{E}_{\mathbb{M}'} $ \\[3pt]
    {aggregated cost}  & $ \displaystyle \boldsymbol{c} = \sum_{m \in \mathbb{M}'} c_{m}$
\end{tabular}\\[4pt]
The performance $p_j^{\text{prov}}$ one module provides to the covered regions $r_j$ can be modeled uniform.
The provided aggregated performance $\boldsymbol{p}_j^{\text{prov}}$ is expected to differ between regions, as contributing modules might provide different coverage.
\begin{figure}[!t]
    \centering
    \includegraphics[width=.9\linewidth]{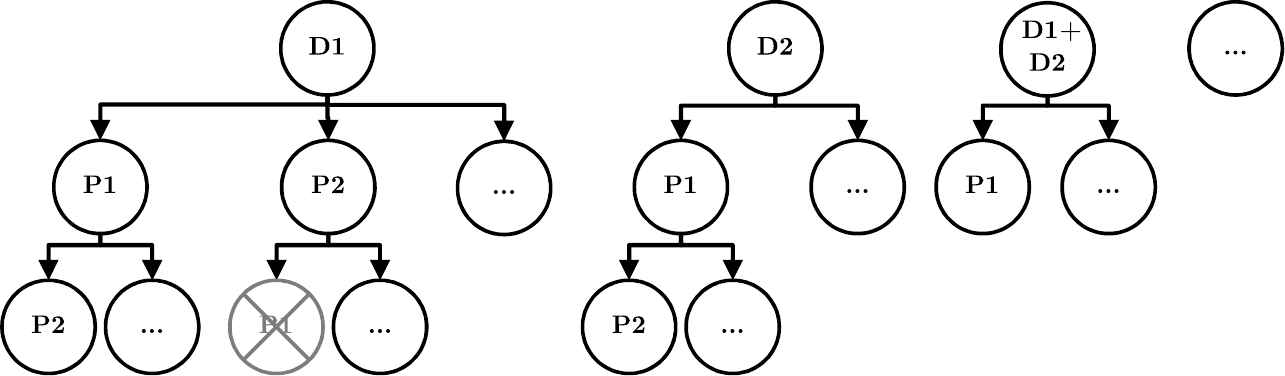}
    \caption{Impression on the configuration tree, depicting \textit{source} modules DX and \textit{non-source} modules PY. Each node represents a full module configuration candidate $\mathbb{M}'$ including all modules on the path from the root to this node. Crossed out elements correspond to invalid or permuted candidates that are pruned or already avoided during tree construction.}
    \label{fig:concept:configTree}
\end{figure}

We propose to construct the set of combination trees with all valid combinations between \textit{source} modules as root nodes and use \textit{non-source} modules in any efficient manner for further tree construction.
The functional architecture of the perception system is fixed w.r.t. the linkability of modules. 
Hence, the order within $\mathbb{M}'$ is irrelevant and it should be ensured that no module permutations exist within the set. 
This can be done either already during tree construction or by pruning afterwards.
Constructing the set of configuration trees with only \textit{source} modules at roots, the \textit{coverage requirement} 
\begin{equation}
    \mathbb{E}_r(\mathbf{s}) \subseteq \mathbb{E}_{\mathbb{M}'}
\end{equation} 
is tested at the top-level of each tree. 
By invalidating trees within the set that do not provide sufficient coverage, the search space for the optimal configuration is reduced.
Further, dynamic constrains from the module relations $R$ as well as from situation dependent requirements invalidate non-compliant nodes, i.e., combination candidates.
Figure~\ref{fig:concept:configTree} shows an exemplary set of configuration trees. 
Each node represents a full module configuration candidate $\mathbb{M}'$ including all modules on the path from the root to this node.
Invalid and redundant combinations are pruned. The empty set ($\mathbb{M}' = \emptyset$) is omitted.

Finally, the \textit{performance requirement}
\begin{equation}
\label{eq:concept:perf}
    p_j^{\text{req}} \leq \boldsymbol{p}_j^{\text{prov}}  ~\forall~ r_j \in \mathbb{E}_r
\end{equation} 
is tested for remaining valid nodes within valid trees. $\mathbb{M}^\ast$ is found by searching through the trees using the optimization criterion \eqref{eq:intro:RC_min}. 
The configuration operator $\text{conf}\left\{\cdot\right\}$ is hence realized by first invalidating insufficient combination candidates within the set of trees using coverage requirement and dynamic constrains, and then by searching for the optimum within the remaining nodes. 
The identified $\mathbb{M}^\ast$ is used for the module reconfiguration during runtime according to the underlying software architecture.

\subsection{Attention Map Application}
\label{sec:concept:appl}
After deployment of the optimal module configuration $\mathbb{M}^{\ast}$ using $\text{conf}\{\cdot\}$, each of the modules applies the MLAM internally. 
Any implementation employing attention based adaptations, like the ones exemplary discussed in Section~\ref{sec:intro}, might be considered. 
These intra-module implementations might employ additional, individual thresholds for the performance requirement of the provided MLAM. In this way, a cost-intensive module might process only data of regions above such thresholds, increasing the potential in reduced resource consumption. As the performance requirement \eqref{eq:concept:perf} needs to remain satisfied, the thresholds need to be set in coordination between the configured modules. A detailed analysis depends on the specific modules and their properties and, therefore, is outside of the scope of this conceptual work.

For the sake of simplicity and without loss of generality, we assume the MLAM to be compatible with the module implementations in this paper. 
In practice, this can be ensured by an additional transformation module.  
Section~\ref{sec:proof} provides a straight-forward example for the attention map application.

\subsection{Concept Summary and Integration}
\begin{figure}[tb]
    \centering
    \includegraphics[width=.75\linewidth]{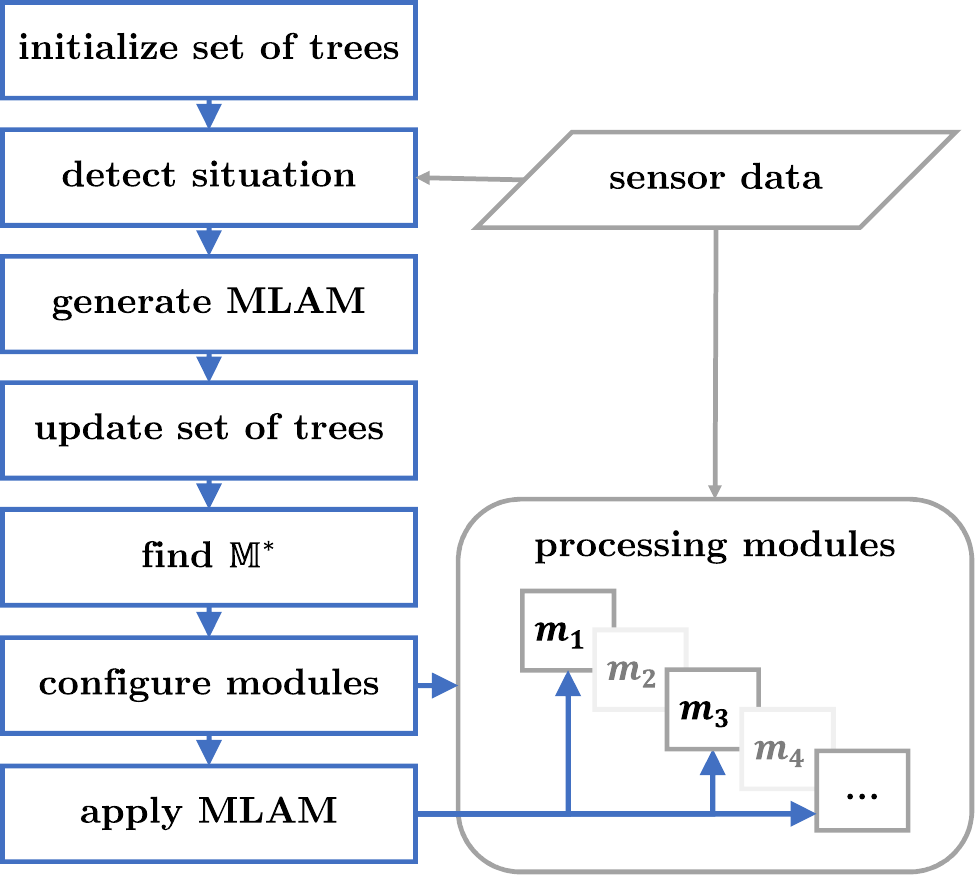}
    \caption{Block diagram of the functional interaction between awareness processing (blue) and an existing perception processing chain (gray). Inactive modules are indicated by fading.}
    \label{fig:concept:ap_blockdiagram}
\end{figure}
Our concept can be split into an initialisation phase and the processing during runtime. Figure~\ref{fig:concept:ap_blockdiagram} outlines the structure of the integration into an existing processing chain. 
First, the set of configuration trees is initialized using the  attributes of the functional modules. These attributes are set beforehand with respect to the actual system at hand. 
During runtime, the current situation is detected from latest sensor data. 
The MLAM is generated and used to update the set of configuration trees.
The best-fitting configuration $\mathbb{M}^{\ast}$ is derived from this updated set and the corresponding processing modules are configured. 
Finally, intra-module attention processes are applied for active modules using the MLAM. 

\section{Proof of Concept}
\label{sec:proof}
In the previous section, we have introduced our generic, modular, and scalable concept for situation-aware environment perception.
In this section, we outline the benefits that can already be achieved by a straight-forward implementation of the approach.
We want to emphasize that we do not strive to provide an all-encompassing solution of the presented concept, but instead to verify the feasibility of its application. 
As no comparably flexible architecture for situation-awareness exists, the presented results are to be seen more as a standalone verification rather than a comparable benchmark.
As exemplary system, we use our automated test vehicle, see \cite{Buchholz2021} for a detailed description.
It is equipped with a range of sensors, comprising lidar, radar, differential GPS, camera, and an inertial measurement sensor. 

\begin{table}[t!]
\centering
\caption{Module overview, presenting the attributes of two model-based object detector (OD) modules and two processing modules. Nominal values for performance are chosen using expert knowledge. Nominal values for cost are based on average processing time.}
\setlength{\tabcolsep}{2pt}
\begin{tabular}{rllccccl}
                                                 & Algorithm               & & Cost               &                     \multicolumn{3}{c}{Performance}          & Relations                 \\
                                                 &                         & &                    &\textit{highway}      &\textit{rural}      &\textit{urban}    & \\
\hline& \\[\dimexpr-\normalbaselineskip+4pt]
\multirow{2}{*}{\rotatebox[origin=c]{90}{source}}&{LIDAR OD}& &{1.0}&{1.0}&{1.0}&{1.0}&  \\
                                                 &{RADAR OD}& &{0.33}&{1.0}&{1.0}&{0.5}&  {low VRU performance}          \\
                                                 &                         & &                    &                    &                    &                    & \\[\dimexpr-\normalbaselineskip+6pt]                            
\hline & \\[\dimexpr-\normalbaselineskip+2pt]
\multirow{5}{*}{\rotatebox[origin=c]{90}{non-source}}&\multirow{2}{*}{TRACKING}&A& 0.1               & 1.0                &  1.0               &  1.0               & requires one OD module \\
                                                 &                         &B& 0.05                & 1.0                &  0.1               &  0.1               & requires one OD module, \\
                                                 &                         & &                    &                    &                    &                    & longitudinal distance only\\
                                                 &                         & &                    &                    &                    &                    & follow drive only\\
                                                 &PLAUSIB                  & & 0.05                & 1.0                &  1.0               &  1.0               & requires TRACKING     \\
\end{tabular}\label{tab:proof:modules}
\end{table}
The remainder of this section outlines the implementation of the concept according to the three steps introduced in Section~\ref{sec:concept} and provides results in two ways: 
In Section~\ref{sec:proof:scenarios}, two artificially constructed scenarios will be evaluated w.r.t. the configuration behavior to gain qualitative insight on the features the concept provides.
In Section~\ref{sec:proof:realWorld}, quantitative results based on real-world data validate the promising capabilities for a reduction in resource consumption.
The provided results correspond to a relative comparison with the \SI[mode=text]{360}{\degree} approach, referred to as naive baseline. 

\subsection{Functional System Architecture}
For the sake of simplicity, we focus our experiments on modules related to model-based object detection. 
The relevant modules $m \in \mathbb{M}$ are summarized in Table~\ref{tab:proof:modules}. 
For brevity, within this proof of concept, we simplify  the aspect of coverage and assume that all modules cover the entire environment, i.e. $\mathbb{E}_m = \mathbb{E}$.
The performance values are derived from expert knowledge, while the cost values are defined as constant and based on average module processing time. 
They are derived from the quantitative results in Section~\ref{sec:proof:realWorld}.
Besides, the module relations are defining module dependencies and situation dependent limitations. 

The functional architecture dynamically connects the modules, so that the planning block can operate with any valid module combination. 
Generally, the lidar object detection module, based on~\cite{Lang2018}, and the radar object detection module~\cite{danzer2019} are connected to the object tracking module, where detections are stabilized.  
Both object detection modules are based on point-cloud-based deep learning networks. 
For tracking, two variants exist: 
Variant A represents a labeled multi-Bernoulli Filter~\cite{Reuter2014}. 
Variant B represents a simplified single object longitudinal distance tracking.
If required, the output of the tracking module, independent of its variant, is fed into the object plausibilization module~\cite{gies2020}. 
Depending on the active configuration, either object detections, tracked objects, or plausibilized objects are fed forward to the planning block.  

\subsection{Situation Detection and Attention Map Generation}
For our proof of concept, the set of possible situations 
\begin{equation}
\label{eq:proof:geolocation}
\mathbb{S} = \left\{\text{highway},\text{rural}, \text{urban}\right\}
\end{equation}
is limited to the geographic location.
A high precision map of the driveable area in a lanelet2~\cite{poggenhans2018} format is available. 
Using GPS information, the situation is extracted from the map.
For the constructed scenarios, one additional influence per scenario will be considered in Section~\ref{sec:proof:scenarios}. 

To derive the separation of the environment into relevant and non-relevant regions, we represent the environment $\mathbb{E}$ as a Cartesian grid of size $m \times m$ with $m=151$, which is centered at the ego vehicle and has a resolution of \SI[mode=text]{1.0}{\meter} per cell.
The Cartesian description is especially well suited for visual interpretation. 
However, since our modules relate to physical sensors with a field of view in polar coordinates, i.e., angular field of view and detection range, the Cartesian attention map is transformed into a polar representation with a \SI[mode=text]{1}{\degree} resolution. Each segment contains a performance requirement $\hat{p}^{\text{req}}$ and a range requirement $\hat{d}^{\text{req}}$, corresponding to the highest performance requirement and the furthest relevant cell within the segment. 
All configuration requirements according to Section~\ref{sec:concept:config} are tested against the polar description.

The set of attention layers $\mathbb{L}$ corresponds to three independent measures for relevance and, therefore, contains three layer functions. 
Table~\ref{tab:proof:sitLink} defines their situational activation.
\begin{table}
\centering
\caption{Activation of the attention layers for the example based on the situation.}
\setlength{\tabcolsep}{4pt}
\begin{tabular}{r|cccc}
\label{tab:proof:sitLink}
\multirow{2}{*}{situation}  & \multicolumn{2}{c}{lane(s)}   & \multirow{2}{*}{ego path} & \multirow{2}{*}{object(s)} \\
                            &  own      & other             &                           & \\
\hline
highway                     & x         &                   & x                         &           \\
rural                       & x         &                   & x                         & x         \\
urban                       & x         & x                 & x                         & x        
\end{tabular}
\end{table}

\textit{Lane} queries surrounding lanes from the available map and projects them onto the attention map. 
Using the lanelet2 API, routing graphs ensure that valid lane transitions are considered. 
For situation dependent granularity, the lanes are split into the own lane of the ego vehicle including valid transitions and all remaining lanes. 
Both lane types can be activated independently.
Their projection results are merged into one layer before layer combination as they correspond to the same measure of relevance.

\textit{Ego Path} defines relevance based on an ego velocity dependent constant turn rate and constant velocity model prediction~\cite{Schubert2008} of the vehicle's position. 
The prediction horizon is set to \SI[mode=text]{3}{\second} and an overly safe uncertainty of the turn rate of \SI[mode=text]{10}{\degree/\second} is used.
In this way, changes in turn rate within the prediction horizon can be represented without employing a more complex model. 
We do not use the planned trajectory within this proof, as it corresponds to lane segments already considered in the \textit{lane} layer.

\textit{Object}  
establishes an awareness for traffic participants that might collide with the ego vehicle. 
It is a counterpart to the naive assumption that relevant traffic participants will only exist on the known, driveable area.
For simplicity of our example, any detected object that lies within a range threshold of $d=15m$ is projected onto the attention map. In case the lidar OD module is active, its detections are used preferably, and detections of the radar OD module otherwise.

For all layers, each intersected cell $j$ is assigned a performance requirement of $p^{\text{req}}_j=1$, as indicated in \figurename~\ref{fig:intro:layerExample:concept}.
Hence, ${p}^{\text{req}}_j$ is limited within $\left[0,1,2,3\right]$, and a cell is deemed relevant above $\theta_{\text{rel}}=0$. 
However, the layer design in combination with the chosen grid size is prone to quantization issues, as the established awareness originates from point-like descriptions, e.g., from nodes of lane center lines or object center positions.
To counteract these issues, the individually created attention maps are dilated using a $5\times 5$ image processing square dilation kernel.
With this dilation, relevance of one cell is inflated to its two neighboring rings and we also establish an increased detection safety, as attention is assigned to more regions.
Figure~\ref{fig:intro:layerExample:realworld} shows an extract of the resulting attention map.

\subsection{Processing Chain Configuration}
\label{sec:proof:mlam}
\begin{figure}[!t]
    \centering
    \includegraphics[width=.8\linewidth]{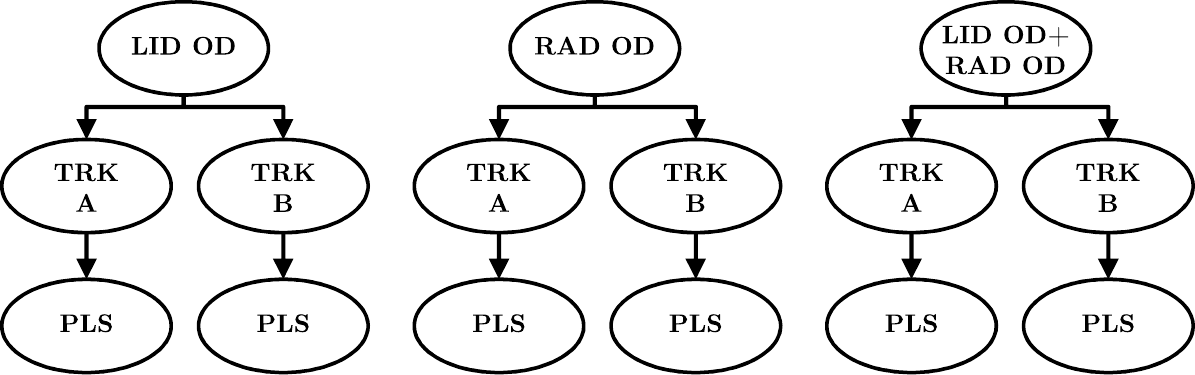}
    \caption{Set of configuration trees based on the modules from Table~\ref{tab:proof:modules}.  Each node represents a full module configuration candidate $\mathbb{M}'$ including all modules on the path from the root to this node. The following abbreviations are used: LIDAR (LID), RADAR (RAD), TRACKING (TRK), PLAUSIB (PLS).}
    \label{fig:proof:config_tree}
\end{figure}

Considering the modules and their relations from Table~\ref{tab:proof:modules}, the set of configuration trees results in 15 nodes including the 3 tree roots. 
The set is visualized in \figurename~\ref{fig:proof:config_tree}.
Again, each node represents a full module configuration including all modules on the path from the root to this node.
The avoidance of invalid configurations and permutations is realized via pruning after construction. Module requirements during runtime, e.g., requiring a gesture detection at a police-officer controlled intersection, are implemented using whitelists and blacklists during the search for $\mathbb{M}^\ast$.
To allow for changing attributes of a module during runtime, the current situation is fed into each active node. Module attributes are thus updated before searching for the optimum.

\subsection{Attention Map Application}
For the sake of simplicity within our proof of concept, the MLAM is applied as a binary data filter, i.e. data are only processed within the active modules for regions with $p_i^{\text{req}}>0$.
Additional intra-module thresholds for $p_i^{\text{req}}$ are not considered.

\subsection{Constructed Scenarios}
\label{sec:proof:scenarios}
With the constructed scenarios we outline a selection of key features that awareness processing provides and show that the connections between attention layers, situations, and modules lead to appropriate and understandable system configurations. 

\begin{figure}[ht]
    \centering
    \includegraphics[width=.8\linewidth]{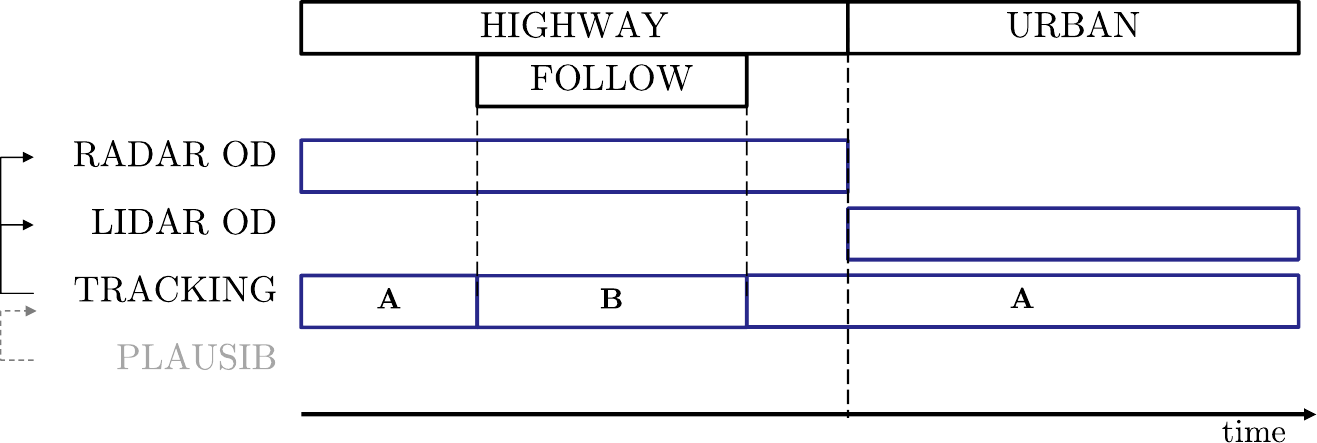}
    \caption{Module configuration for the highway scenario with follow drive and a following urban scenario without relevant objects. {A}/{B} represent algorithm variants, a blue bar represents the active state of a module, arrows represent the module dependencies.}
    \label{fig:proof:CS_highway}
\end{figure}

In \figurename~\ref{fig:proof:CS_highway}, a transfer situation between a highway and an urban location is shown. On the highway, the \textit{ego path} and the \textit{own lane} attention layer are active, resulting in a maximum possible performance requirement of $2$. 
The radar OD module is employed due to its lower cost compared to the lidar OD module. 
To meet the performance requirement, tracking variant A is employed in addition.
A follow drive sub-situation is detected for a short duration on the highway. 
During this time, the cost-efficient tracking variant B is chosen. This reflects the expert knowledge in the modeling of relations (cf. Table~\ref{tab:proof:modules}), which considers longitudinal distance tracking sufficient for following tasks.
Transferring into an urban situation, the performance of the radar OD module degrades to $0.5$, reflecting its low performance in vulnerable road user (VRU) detection. 
At the same time, the \textit{object} layer is activated in addition to the \textit{ego path} and \textit{lane} layer. This results in a maximum performance requirement value of $3$. 
However, for this scenario, we assume that no detection becomes relevant for the \textit{object} layer within the shown duration.
Consequently, its performance requirement equals 0 for all regions. Thus, the maximum performance requirement to be met is $2$, as \textit{ego path} and \textit{lane} partially overlap. The lidar OD module together with the tracking variant A satisfies the requirement at lowest cost.

\begin{figure}[ht]
    \centering
    \includegraphics[width=.8\linewidth]{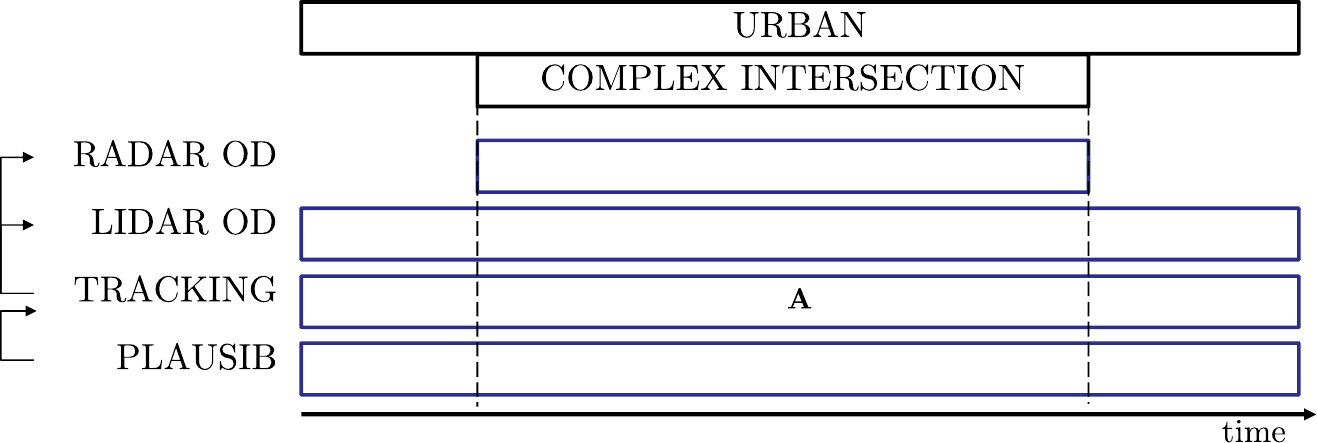}
    \caption{Module configuration for the urban scenario with the complex intersection and with relevant objects. {A}/{B} represent algorithm variants, a blue bar represents the active state of a module, arrows represent the module dependencies.}
    \label{fig:proof:CS_urban_intersection}
\end{figure}

In \figurename~\ref{fig:proof:CS_urban_intersection}, a complex intersection is detected in an urban location. 
This might correspond to faulty traffic lights, a traffic jam on the intersection, roadworks, or to other conditions. 
The attention layers \textit{ego path}, \textit{lane}, and \textit{object} are active. Since  we assume that relevant objects are detected in this scenario, this results in an overall maximum performance requirement of $3$.
Outside of the sub-situation of a complex intersection, the lidar OD module, tracking variant A, and the plausibilization module are employed, satisfying the requirement at lowest cost. 
Once the complex sub-situation is detected, a dynamic requirement for a second OD module is added. 
Consequently, the radar OD module is activated. The configuration is then identical to the vehicle setup of the naive baseline.

The presented scenarios show that inter-dependencies between modules as well as module dependencies that relate to the situation provide great flexibility for a diverse range of systems for concept application.
Similarly, the situation dependent activation of attention layers allows for a general as well as a specific definition of relevant regions.
Reestablishing the naive baseline setup remains possible, if required.

\subsection{Real-World Example}
\label{sec:proof:realWorld}

\begin{figure}[!t]
    \centering
    \includegraphics[width=.65\linewidth]{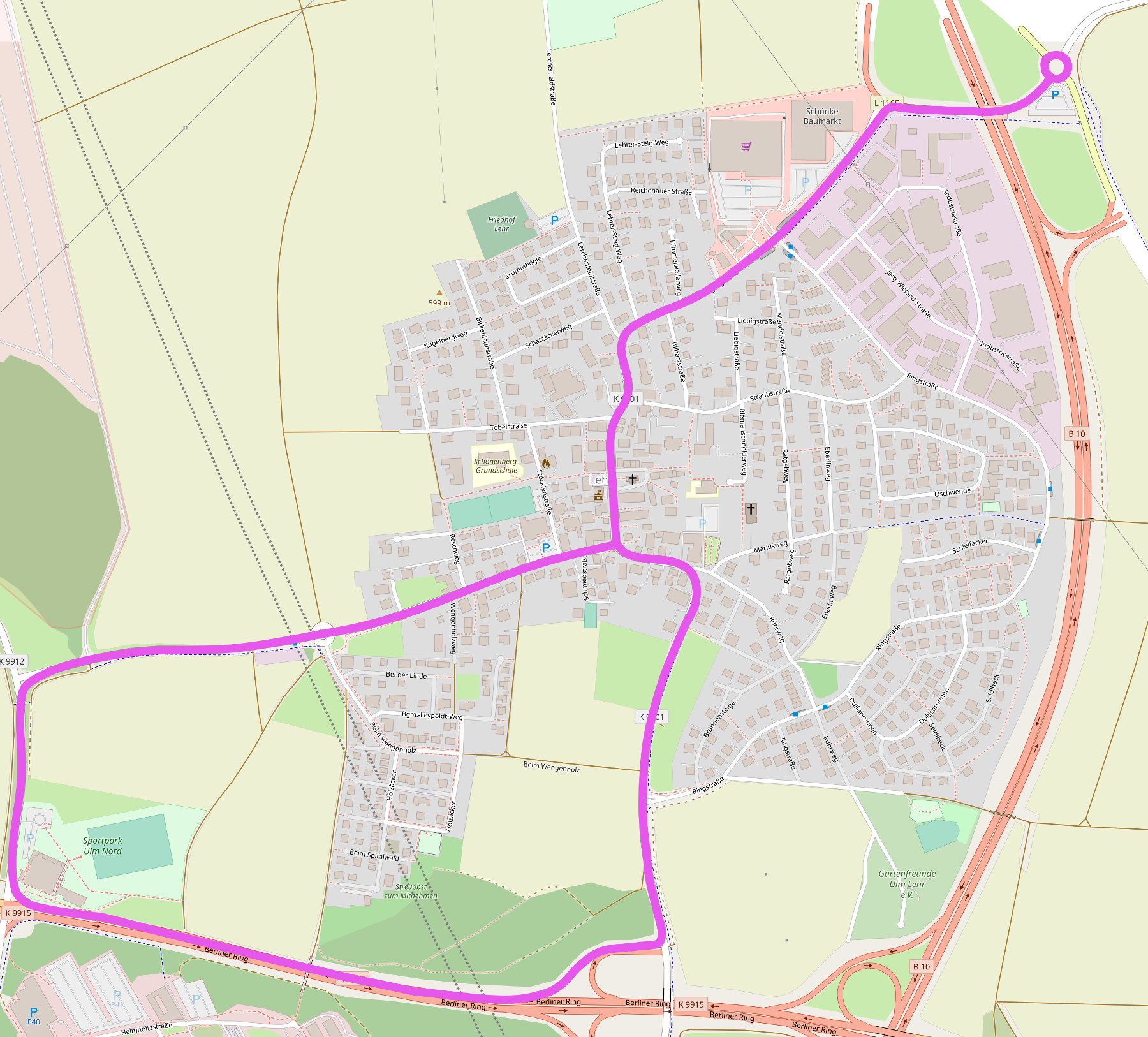}
    \caption{Outline of the route for the quantitative results in pink (map data from OpenStreetMap~\cite{OpenStreetMap}).
    }
    \label{fig:proof:lehr_runde}
\end{figure}

The real-world results are derived from pre-recorded data taken in the vicinity of the Ulm University in Ulm, Germany.
Simulation in a post-processing manner is conducted on a consumer PC equipped with an Intel i7-6850 CPU and a Nvidia TitanX 12GB GPU on 64GB RAM. 
The driven route contains various aspects of urban and rural driving with intersections, roundabouts, merging/splitting lanes, etc., and a short highway-like section, as outlined in \figurename~\ref{fig:proof:lehr_runde}. 
Overall, the data comprise a track length of approximately \SI[mode=text]{5}{\kilo\meter}.

\subsubsection{Quantitative Results}
\begin{figure}[!t]
    \centering
    \includegraphics[width=.65\linewidth]{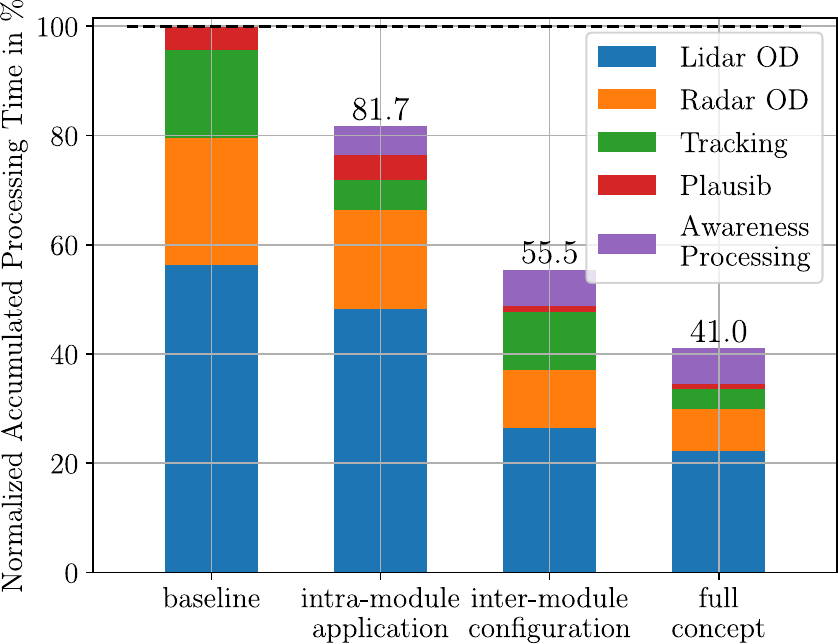}
    \caption{Accumulated processing time per module, normalized w.r.t. the naive baseline.}
    \label{fig:proof:accTime}
\end{figure}
The straight-forward implementation of our concept from the first parts of this section was applied to the dataset described above. 
To gain insight on the potential of awareness processing, \figurename~\ref{fig:proof:accTime} presents the normalized accumulated processing time over all modules and the computational overhead induced by awareness processing in comparison to the naive baseline. 
On the left side, the approximated cost values from Table~\ref{tab:proof:modules} are reflected. 
The two center columns show the individual application of the two key aspects of awareness processing, i.e., 
processing only relevant data within active modules and employing only a subset of active modules.
Due to the nature of the deep learning based detection modules the reduction in processing time for the detection modules is much lower than for the tracking module, where processing of irrelevant or clutter detections can be avoided. 
Considering only a small number of available modules in our example, employing a subset of these provides larger reductions than processing only relevant data within all modules. Here, the tracking module requires similar processing time as in the baseline. Processing time of all other modules is reduced significantly.
The most right column shows the full concept application.
Combining our key aspects, an impressive relative reduction of \SI[mode=text]{59.0}{\percent} can be achieved. 
The induced overhead of awareness processing, which is implemented using solely CPU resources within our example, accumulates to \SI[mode=text]{6.5}{\percent} of the naive baseline processing time. 
This overhead is neatly put into perspective against the potential that awareness processing provides. 
For the sake of brevity, only the full concept application is investigated further.

\begin{figure}[t!]
    \centering
    \subfloat[Average number of active modules over geographic location.]{
        \includegraphics[width=.75\linewidth]{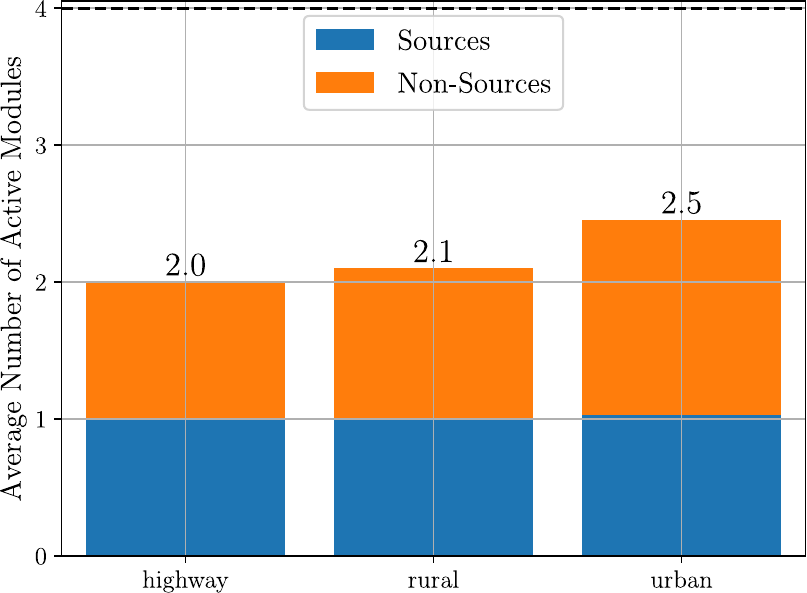}
        \label{fig:proof:activeConfig:sit}}
    \hfill
    \subfloat[Relative module uptime]{
        \includegraphics[width=.75\linewidth]{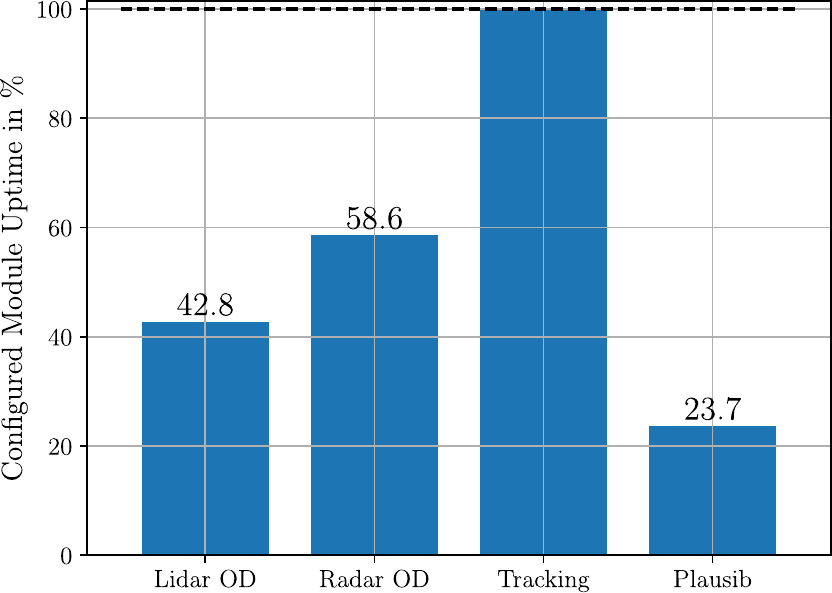}
        \label{fig:proof:activeConfig:module}}
    \caption{Active module configuration compared to the naive baseline (4 modules at \SI[mode=text]{100}{\percent} uptime).}
    \label{fig:proof:activeConfig}
\end{figure}

Figure~\ref{fig:proof:activeConfig} shows an overview of the active modules throughout the test sequence in comparison to the naive baseline (dashed line), which would apply all four modules continuously. 
In \figurename~\ref{fig:proof:activeConfig:sit}, a significant reduction in average active modules over situation can be seen. 
The variation between the different situations reflects the attention layer activation from Table~\ref{tab:proof:sitLink}. 
As expected, the urban scenario receives the lowest reduction due to the additional aggregated attention towards oncoming objects, which become relevant within the \textit{object} layer, and the attention towards their lanes from the \textit{lane} layer.
Figure~\ref{fig:proof:activeConfig:module} shows the relative uptime for the individual modules. 
Here, cost and performance from Table~\ref{tab:proof:modules} as well as module relations are reflected. 
Even with its low cost, the plausibilization module is employed least often due to its requirement for both an object detection module and a tracking module to be active. 
With low cost and high performance, one tracking variant is active at all times. 
Besides, the more expensive lidar OD module is employed less often than the radar OD module. 
It is notable that the uptime of the detector modules sum to over \SI[mode=text]{100}{\percent}, i.e., they are running jointly in some cases.
This corresponds to the constructed intersection scenario from Section~\ref{sec:proof:scenarios}.

\begin{figure}[t!]
    \centering
    \subfloat[Processing time for the awareness processing and the modules. Whiskers correspond to $3\times\text{inter-quartile range}$.]{
        \includegraphics[width=.9\linewidth]{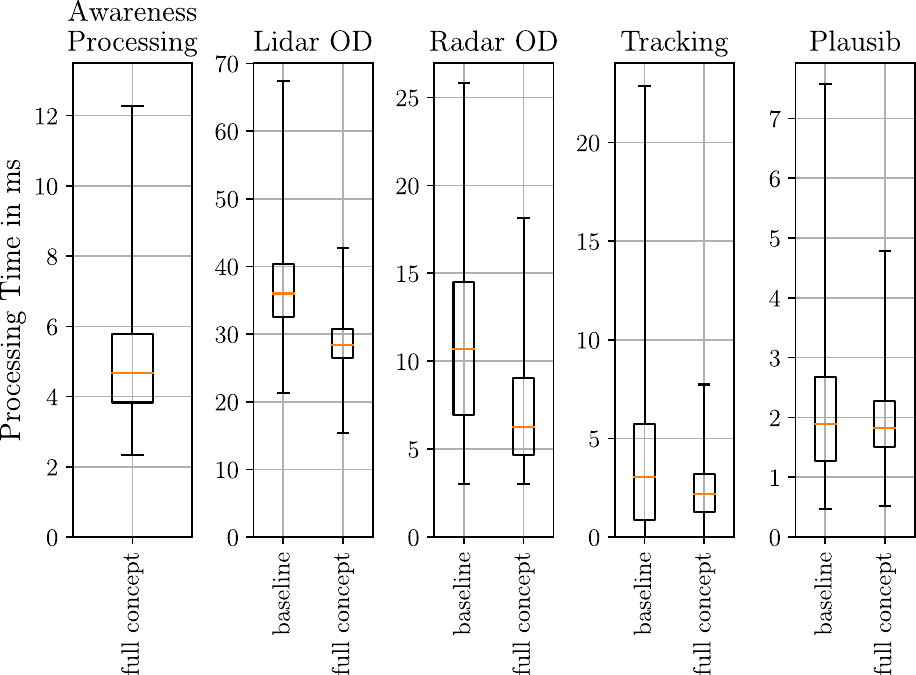}
        \label{fig:proof:module_box}}
    \hfill
    \subfloat[Hardware load comparison. Whiskers correspond to $3\times\text{inter-quartile range}$.]{
        \includegraphics[width=.75\linewidth]{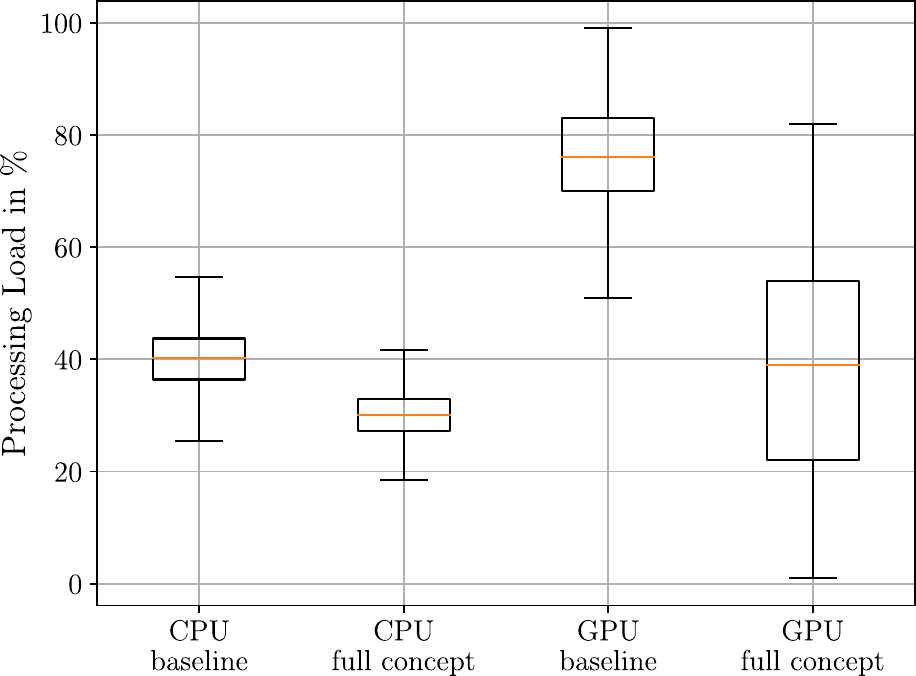}
        \label{fig:proof:usage}}
    \caption{Comparison between the awareness processing and the naive baseline.}
    \label{fig:proof:comparison}
\end{figure}

Comparative results for the module processing times after application of the identified MLAM are shown in \figurename~\ref{fig:proof:module_box} as boxplots. 
The boxplots present the processing time in every active cycle, where the whiskers correspond to $3\times\text{inter-quartile range (IQR)}$, and the orange lines to the median of the processing time. 
Although the given values are system dependent, they clearly indicate the effectiveness of the concept.
The computational overhead that is required for awareness processing is shown on the far left side.  
The remaining plots show direct comparisons of the processing times between the naive baseline and our full concept application. 
For all modules, two key observations can be drawn: 
First, the median of the processing time is reduced, as was our expectation. 
Second, the spread of the IQR is greatly reduced as well. 
This is a crucial verification of the effectiveness of our approach, as it shows that a significant part of processing resources can be freed by enforcing resource allocation towards relevant areas. 
Taking the tracking module as an example, this behavior can also be explained situation dependent: 
when passing by a mostly occupied parking lot that does not have an exit to the current road of the vehicle, a large amount of object detections will have to be processed in the naive baseline. 
These detections are not relevant to the driving task, but require a large amount of resources that can be saved already on object detection module level. 
Hence, reductions in resource consumption are larger the earlier the processing can be reduced.
In addition, \figurename~\ref{fig:proof:module_box} verifies our hypothesis from Section~\ref{sec:problem:assumption}: the induced processing overhead is non-negligible, but easily compensated by the resources freed within the affected modules.

We acknowledge that a reduction in the processing time of a module does not necessarily correspond to a lower energy intake. Therefore, \figurename~\ref{fig:proof:usage} evaluates above observations on hardware level, showing the loads of CPU and GPU, which are tightly coupled with the energy intake.
For both CPU load and GPU load, the median is reduced. 
Besides, the module uptime from \figurename~\ref{fig:proof:activeConfig:module} is reflected in the significant downward spread of the GPU load. Thus, from this results, it can be concluded that despite the additional calculations for our situation-awareness module, the energy consumption is decreased.

Summarizing the presented results of the real-world application, we conclude that a significant reduction in resource consumption is achieved already with a straight-forward implementation of awareness processing and that the situational activation of attention layers and the parameterization of the module's attributes interact as we have intended.  
Using the MLAM to identify the optimal subset of active modules as well as to enforce data processing only to relevant regions, the processing time and the hardware load decreased significantly. 
Applying our concept for awareness processing to our automated vehicle the results on planning level remained plausible w.r.t. to our baseline. 
Still, driving performance evaluation is outside of the scope of this work.
Our resulting reductions can be summarized as follows:\\[4pt]
\begin{tabular}{rcl}
    \text{CPU load reduction (median)}: & & \SI[mode=text]{25.4}{\percent} \\
    \text{GPU load reduction (median)}: & & \SI[mode=text]{48.7}{\percent} \\
    \text{Relative overall processing time reduction}: & & \SI[mode=text]{59.0}{\percent}
\end{tabular}

\section{Conclusion}
In this work, we have introduced \textit{awareness processing} as a flexible concept for situation-aware environment perception that is both scalable as well as modular. The key aspects of our concept are 
1)~the definition of situations to be considered and layers to distinguish relevant from non-relevant regions, resulting in a multi-layer attention map (MLAM),
2)~the dynamic optimization of the module configuration that optimally satisfies the requirements of the MLAM, 
and 3)~the enforcement of data processing to only relevant regions defined by the MLAM.
Applications of our concept can be designed as specific as desired for the systems that they are derived for.
Providing this flexibility between general application and specific application for any system at hand represents the core idea of our contribution.

The exemplary application that we have presented provides the proof for the feasibility of the concept application and, most importantly, shows that already straight-forward implementations can lead to significant reduction in resource consumption.
For the provided real-world example in urban, rural, and highway-like conditions, an overall reduction in processing time of \SI[mode=text]{59.0}{\percent} was achieved. 
The reduction is reflected in the reduction of the CPU and the GPU median load by \SI[mode=text]{25.4}{\percent} and \SI[mode=text]{48.7}{\percent}, respectively. 

The quantization of the cost and the performance of a module is a central as well as challenging element for the application of our concept.
While we have presented some options, the research in this field is ongoing, and extensive solutions are yet to be found.
Stellet et al.~\cite{Stellet2020} have recently published a survey on the state of the art in performance assessment for automated driving.
Considering their conclusions, in future work, we aim to integrate our application into a sophisticated simulation framework, e.g., \textit{CARLA}~\cite{Dosovitskiy17}.

We have shown that a reduction in resource consumption is possible where performance requirements can be met. 
Similarly, integration of self-awareness aspects should be pursued in continued concept applications, so that synergies between self-awareness and situation-awareness can manifest.

\ifCLASSOPTIONcaptionsoff
  \newpage
\fi

\begin{IEEEbiography}[{\includegraphics[width=1in,height=1.25in,clip,keepaspectratio]{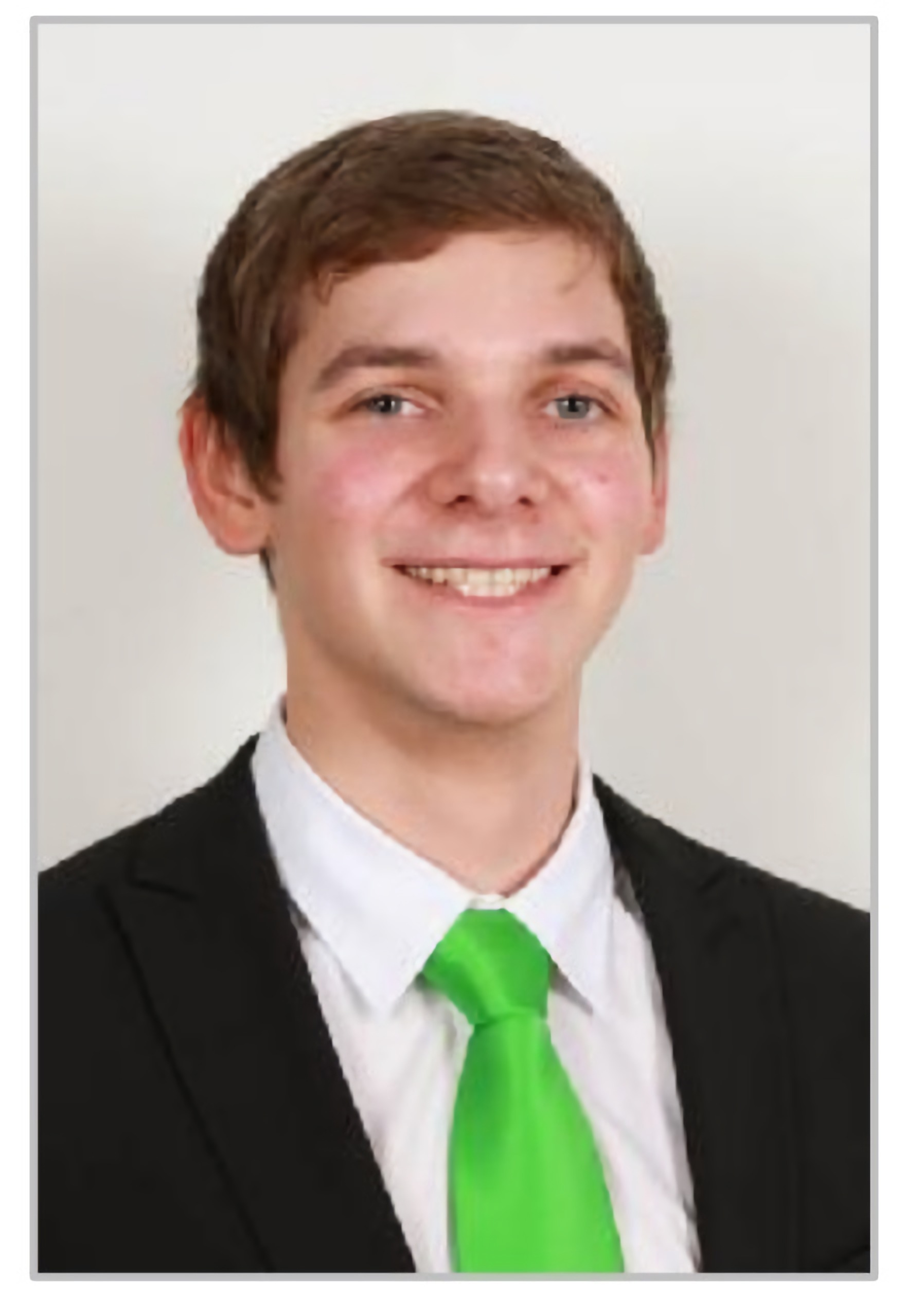}}]{Matti Henning}
(matti.henning@uni-ulm.de) received his bachelor's degree in mechatronics from the Bremen City University of Applied Sciences, Germany, and his master's degree in navigation and field robotics from the Leibniz University Hanover, Germany. After working in the automotive radar series development of ADAS at Continental AG for over three years, he is currently a researcher at the Institute of Measurement, Control, and Microtechnology at Ulm University, working towards his PhD degree in the field of automated driving.
\end{IEEEbiography}

\begin{IEEEbiography}[{\includegraphics[width=1in,height=1.25in,clip,keepaspectratio]{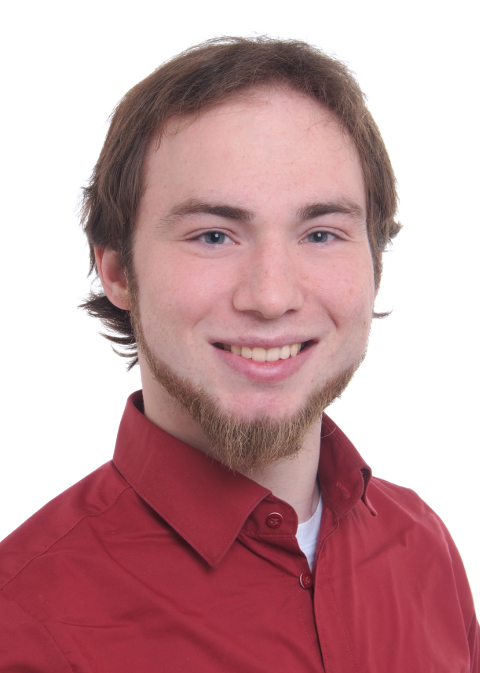}}]{Johannes Müller}
(johannes-christian.mueller@uni-ulm.de, Member, IEEE) received his bachelor's and master's degree in electrical engineering and information technology from Karlsruhe Institute of Technology, Germany, in 2013 and 2016, and his PhD degree from Ulm University, Germany, in 2021, respectively. Since 2017, he has been a researcher at the Institute of Measurement, Control, and Microtechnology at the Ulm University. His research interests are motion planning and reliability estimation for connected automated vehicles.
\end{IEEEbiography}

\begin{IEEEbiography}[{\includegraphics[width=1in,height=1.25in,clip,keepaspectratio]{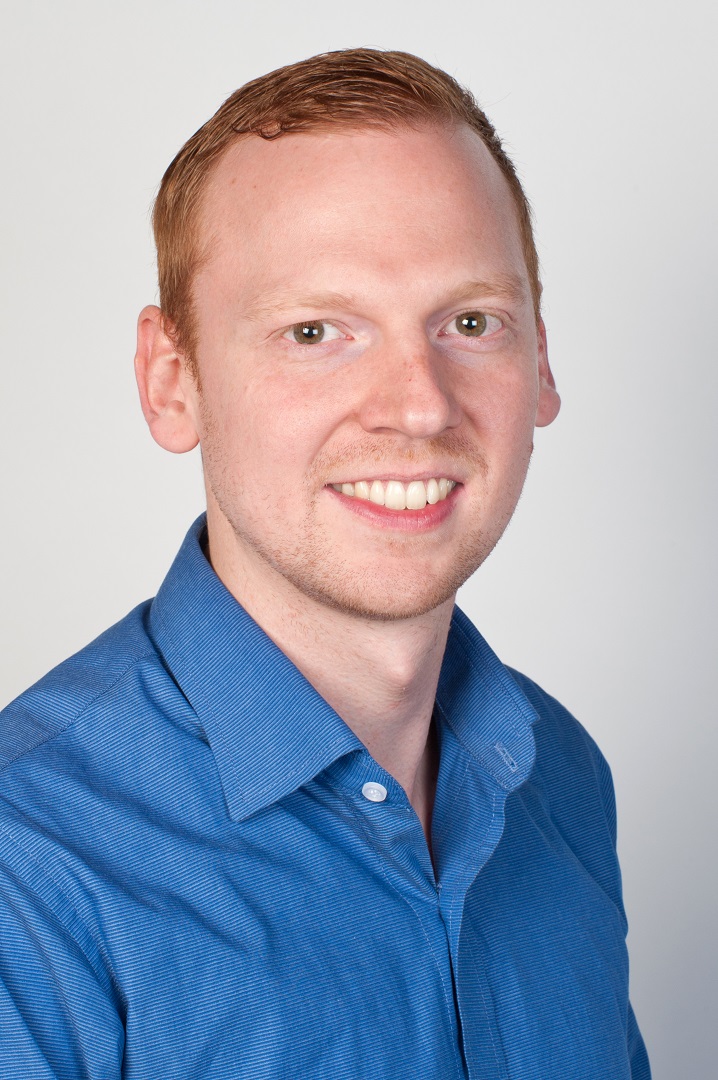}}]{Fabian Gies} 
received his bachelor's degree in electrical engineering from the University of Applied Sciences, Karlsruhe, Germany and his master's degree in communications and computer engineering from the Ulm University, Germany. He joined the Institute of Measurement, Control and Microtechnology at Ulm University, Germany in 2016 as an academic researcher, where he worked towards his Ph.D. in the field of environment perception of automated vehicles including topics of sensor fusion, multi-object tracking, dynamic occupancy grid mapping and track-to-track fusion. 

\end{IEEEbiography}
\begin{IEEEbiography}[{\includegraphics[width=1in,height=1.25in,clip,keepaspectratio]{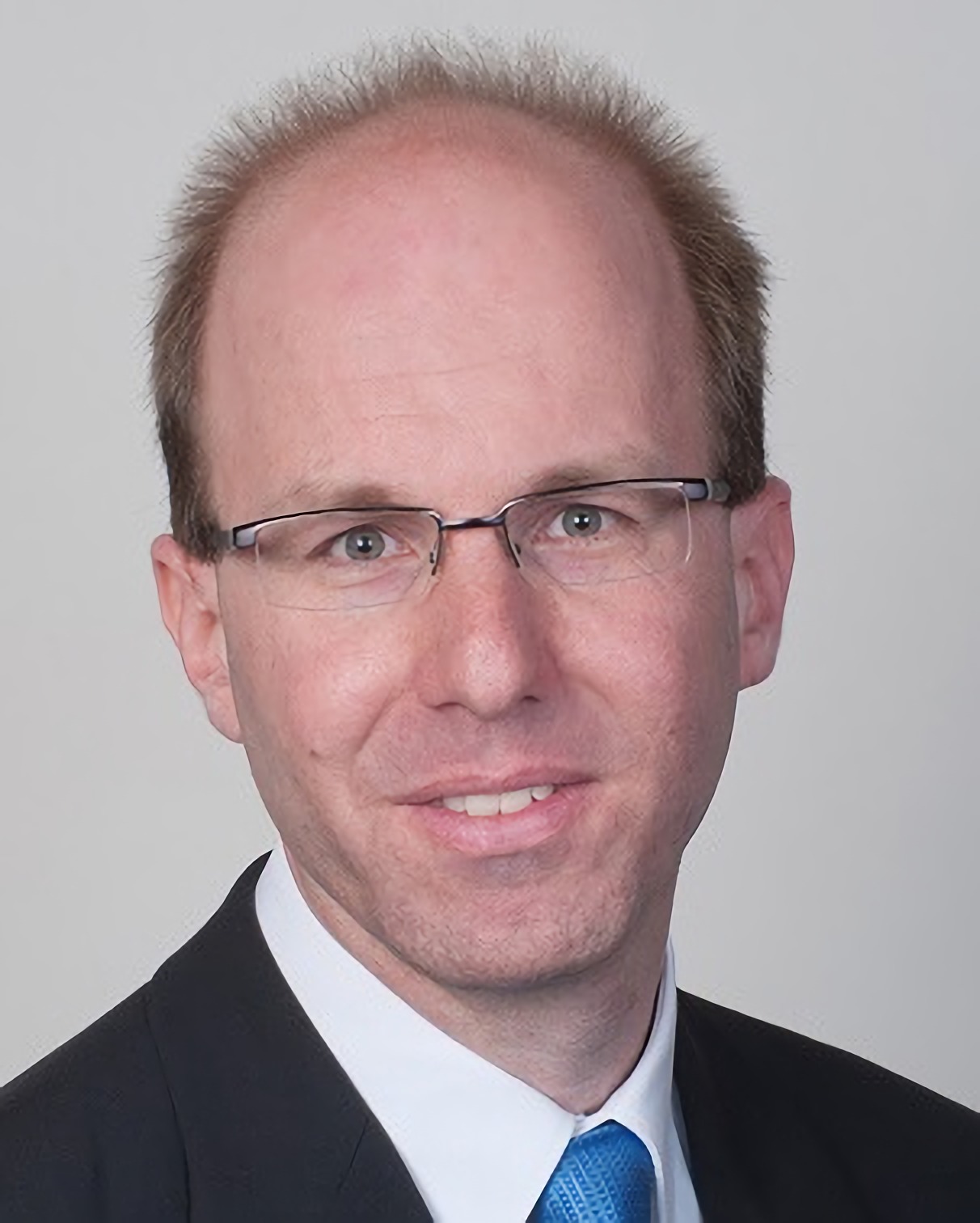}}]{Michael Buchholz}
(michael.buchholz@uni-ulm.de) received his Diploma degree in Electrical Engineering and Information Technology as well as his Ph.D. from the faculty of Electrical Engineering and Information Technology at Karlsruhe Institute of Technology, Germany.  Since 2009, he is serving as a research group leader and lecturer at the Institute of Measurement, Control and Microtechnology at Ulm University, Germany.  His research interests comprise connected automated driving, electric mobility, modelling and control of mechatronic systems, and system identification.
\end{IEEEbiography}

\begin{IEEEbiography}[{\includegraphics[width=1in,height=1.25in,clip,keepaspectratio]{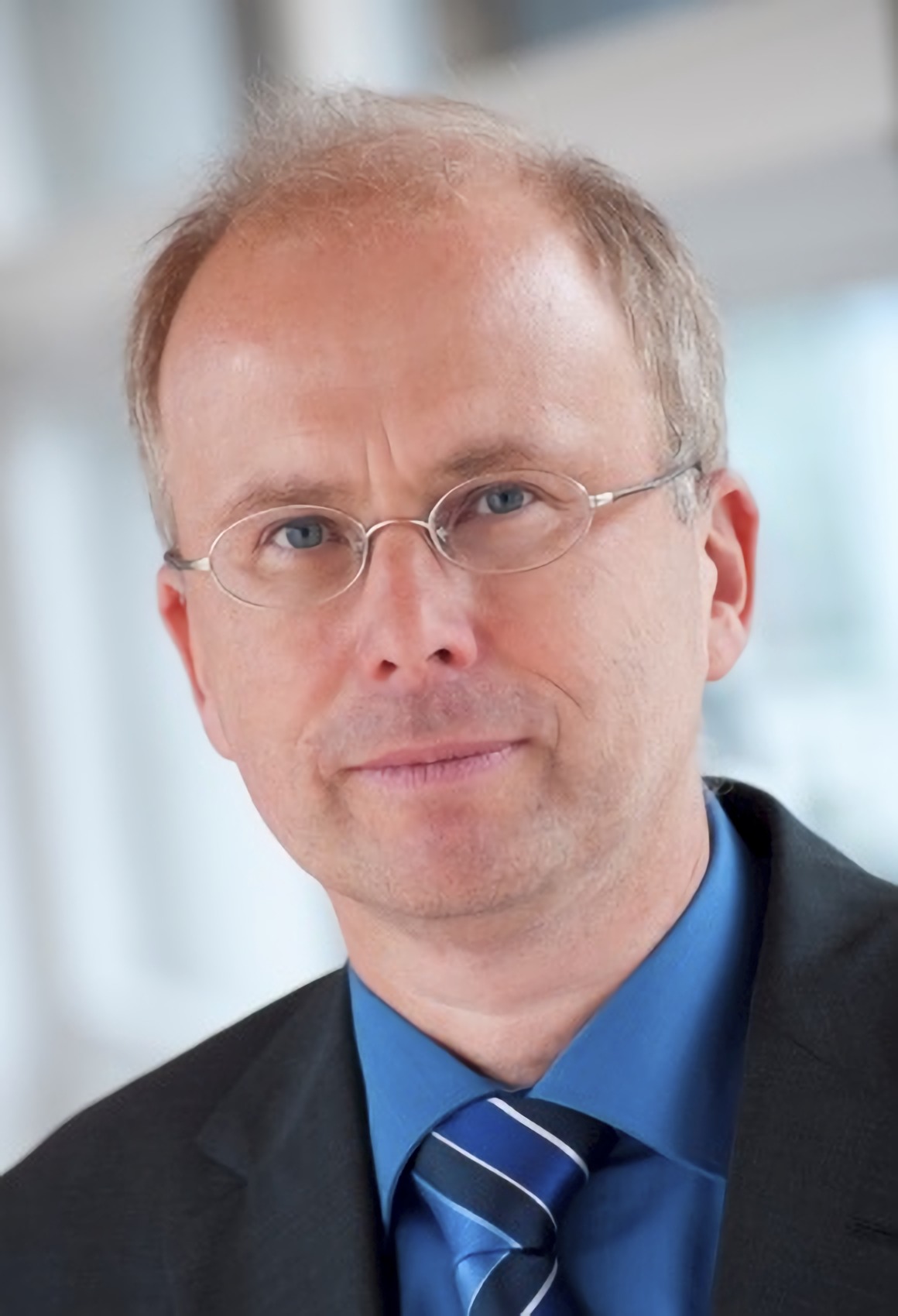}}]{Klaus Dietmayer}
(klaus.dietmayer@uni-ulm.de, IEEE, Member) was born in Celle, Germany in 1962. He received his Diploma degree (equivalent to M.Sc. degree) in 1989 in electrical engineering  from  the  Technical  University  of Braunschweig, Germany, and the Dr.-Ing. degree (equivalent to Ph.D.) in 1994 from the University of Armed Forces in Hamburg, Germany. In 1994, he joined the Philips Semiconductors Systems Laboratory in Hamburg, Germany as a research engineer. Since 1996, he became a manager in the field of networks and sensors for automotive applications. In 2000, he was appointed to a professorship at Ulm University in the field of measurement and control. Currently, he is Full Professor and Director of the Institute of Measurement, Control and Microtechnology in the school of Engineering and Computer Science at Ulm University. His research interests include information fusion, multi-object tracking, environment perception for advanced automotive driver assistance, and E-Mobility. Klaus Dietmayer is member of the IEEE and the German society of engineers VDI/VDE.
\end{IEEEbiography}

\end{document}